\documentclass[12pt]{article}

\usepackage[margin=1in]{geometry}
\usepackage[T1]{fontenc}
\usepackage{CJKutf8, amssymb, amsmath, booktabs, hyperref, orcidlink}

\usepackage[many]{tcolorbox}
\tcbuselibrary{listings}
\newtcblisting{promptbox}[1][]{
    enhanced,
    listing only,
    colback=gray!5!white,
    colframe=gray!75!black,
    fonttitle=\bfseries,
    title=#1,
    boxrule=0.5pt,
    arc=3pt,
    left=4pt,
    right=4pt,
    top=4pt,
    bottom=4pt,
    enlarge top by=0.5\baselineskip,
    enlarge bottom by=0.5\baselineskip,
    breakable,
    listing options={
        basicstyle=\ttfamily\small,
        breaklines=true,
        breakatwhitespace=true,
        breakindent=0pt,
        columns=fullflexible,
        literate={ő}{{\H{o}}}1
                 {é}{{\'e}}1
                 {–}{{\textendash}}1
    }
}

\newtheorem{theorem}{Theorem}

\usepackage[style=trad-unsrt, citestyle=numeric-comp, giveninits=true, doi=false, url=false, eprint=false, isbn=false]{biblatex}

\begin{document}

\title{Autonomous disproofs of the sum-product conjecture over $\mathbb R$ with GPT-5.5 Pro\thanks{Project repository: \url{https://github.com/yichenhuang/sum-product}}}

\author{Yichen Huang (黄溢辰)\orcidlink{0000-0002-8496-9251}\\
Santa Clarita, California 91350, USA\\
\href{mailto:huangtbcmh@gmail.com}{huangtbcmh@gmail.com}}

\begin{CJK}{UTF8}{gbsn}

\maketitle

\end{CJK}

\begin{abstract}

OpenAI's recent disproof of the Erd\H{o}s unit distance conjecture marked a milestone for AI in mathematics. It also inspired another breakthrough: a human disproof of the Erd\H{o}s--Szemer\'edi sum-product conjecture over $\mathbb R$. In this paper, we present a simple agent built on GPT-5.5 Pro. Using a problem-agnostic, three-stage prompting pipeline---proof-plan proposal, proof construction, and review---the agent autonomously generated correct proofs that the sum-product conjecture is false over $\mathbb R$ in 7 of 8 independent trials; in the remaining trial, it identified an unresolved gap in its argument. The seven proofs are diverse: some are close to existing unit-based constructions, while others avoid units by using $L^p$-type regions of algebraic integers. The system used an average of 132.4k reasoning tokens per trial. We release the code, intermediate outputs, and generated proofs, providing a reproducible, data-contamination-free case study in autonomous proof generation.

\end{abstract}

\section{Introduction}

Recently, OpenAI announced that an internal AI model autonomously disproved the Erd\H{o}s unit distance conjecture \cite{OAI26, ABG+26}. This resolved a famous long-standing open problem in discrete geometry and marked a milestone in the application of AI to mathematical research. Inspired by the unit-distance counterexample, Bloom, Sawin, Schildkraut, and Zhelezov \cite{BSSZ26} soon made a second breakthrough: a human disproof of the Erd\H{o}s--Szemer\'edi sum-product conjecture over $\mathbb R$.

\paragraph{The Erd\H{o}s--Szemer\'edi sum-product conjecture.}For a finite set $A$ in a commutative ring, write
\begin{equation}
A+A=\{a+b:a,b\in A\},\qquad A\cdot A=\{ab:a,b\in A\}.
\end{equation}
The sum-product problem asks how small these two sets can be simultaneously. For $A\subset\mathbb Z,\mathbb R,\mathbb C$, it was conjectured that at least one of them must have essentially maximal size:
\begin{equation}
        \max\{|A+A|,\ |A\cdot A|\}\ge |A|^{2-o(1)}.
\end{equation}
This conjecture first appeared in \cite{Erd76}, where Erd\H{o}s stated it with particular emphasis on the integer case. It is usually called the Erd\H{o}s--Szemer\'edi sum-product conjecture because Erd\H{o}s and Szemer\'edi \cite{ES83} proved the first general results in this direction. In the real case, Solymosi proved a $|A|^{4/3-o(1)}$ lower bound \cite{Sol09}; Konyagin and Shkredov broke the $4/3$ barrier \cite{KS16}; the best known lower bound is $|A|^{4/3+10/4407-o(1)}$, due to Cushman \cite{Cus25}.

Bloom, Sawin, Schildkraut, and Zhelezov disproved the sum-product conjecture over $\mathbb R$:
\begin{theorem} [\cite{BSSZ26}] \label{thm:main}
There is an absolute constant $c>0$ and arbitrarily large finite sets $A\subset\mathbb R$ such that
\begin{equation}
        \max\{|A+A|,\ |A\cdot A|\}\le |A|^{2-c}.
\end{equation}
\end{theorem}
Their counterexamples are constructed inside rings of algebraic integers in totally real number fields of arbitrarily large degree and uniformly bounded root discriminant, and are then embedded into $\mathbb R$. Martinet's class-field towers supply the required number-theoretic inputs. The use of large-degree number fields is the essential new source of counterexamples, and may help explain why the construction was invisible to earlier geometric and combinatorial methods.

\paragraph{AI follow-up results.}Levent Alp\"oge \cite{Alp26} announced that an agent based on Anthropic's Claude Mythos autonomously proved Theorem \ref{thm:main}. This proof uses the same basic construction as \cite{BSSZ26}. Anthropic later publicly released Claude Fable 5, officially described as a ``Mythos-class'' model. Nevertheless, the exact Claude Mythos model and the agent used by Alp\"oge remain non-public. Neither the probability that a run produces a correct proof nor the computational cost has been reported.

Sébastien Bubeck \cite{Bub26} published a GPT-5.5 Pro transcript prompted by Boris Alexeev \cite{Ale26}. However, this transcript does not represent an autonomous disproof. A key prompt supplied the model with the unit-distance proof and explicitly instructed it to use the same approach to disprove the sum-product conjecture. This is a substantial mathematical hint. Thus the transcript is best understood as a human-guided test of GPT-5.5 Pro rather than as an autonomous discovery of the sum-product counterexample.

\paragraph{Our contribution.}We use GPT-5.5 Pro as the underlying model in order to obtain a clean experiment free from the risk of data contamination. GPT-5.5 Pro was released before the first public proof \cite{BSSZ26} appeared, and all runs were conducted with web search disabled. Therefore the model could not have retrieved any public proof at inference time, nor could any such proof have appeared in its training data.

In this paper, we present a simple agent built on GPT-5.5 Pro. The agent uses a three-stage prompting pipeline: proof-plan proposal, proof construction, and critical review. In 7 out of 8 independent trials, the agent autonomously proved Theorem \ref{thm:main} correctly. In the remaining trial, it did not complete the proof, but correctly identified the gap rather than presenting an incomplete argument as a finished proof. The experiment was computationally modest, averaging $132.4$k reasoning tokens per trial across all eight runs. Thus, in hindsight, an autonomous disproof of the Erd\H{o}s--Szemer\'edi sum-product conjecture over $\mathbb R$ was well within the capabilities of a public model released before the human proof \cite{BSSZ26}, without relying on complex scaffolding or models that were non-public at the time. We release the agent's source code, all intermediate model outputs, and generated proofs on GitHub at \url{https://github.com/yichenhuang/sum-product}.

The agent also generated proofs that are substantially different from all public proofs. These proofs do not use units at all; instead, they construct counterexamples by selecting algebraic integers from a carefully chosen region and then embedding them into $\mathbb R$. They require less number-theoretic background and are of independent mathematical interest.

\section{Agent}

While termed an agent, our system uses a minimalist architecture: it runs a three-round conversation, prompting the underlying model to sequentially identify a promising approach, construct a rigorous proof, and perform a critical review.

The system prompt provides the context and objective:

\begin{promptbox}[System Prompt]
The provided theorem was recently proved, and the result is now accepted by the mathematical community. Your task is to independently construct a proof.
\end{promptbox}

The first prompt states the theorem and asks the model to identify a promising approach toward proving it:

\begin{promptbox}[First Prompt]
**Theorem.** There exist an absolute constant $\delta > 0$ and an infinite sequence $\{A_i\}$ of finite subsets of $\mathbb{R}$ such that $|A_i| \to \infty$ and $\max(|A_i+A_i|, |A_iA_i|) \le |A_i|^{2-\delta}$ for all $i$, where $A_i+A_i = \{ a+b : a,b \in A_i \}$ and $A_iA_i = \{ ab : a,b \in A_i \}$ are the sumset and product set of $A_i$, respectively.

**Remark.** This result disproves the Erdős–Szemerédi Sum-Product Conjecture over $\mathbb{R}$.

***

As a first step toward a complete proof, identify the most promising approach and develop it as far as possible. Present a step-by-step proof plan. Prove any intermediate claims you can. For each remaining gap, state it precisely and assess whether it can plausibly be closed within the approach or poses a serious obstruction. Include enough detail for the next round to continue from where you leave off.
\end{promptbox}

The second prompt asks the model to construct a complete and rigorous proof:

\begin{promptbox}[Second Prompt]
Construct a complete and rigorous proof, adapting your approach as needed.
\end{promptbox}

The third prompt asks the model to critically review and refine its proof:

\begin{promptbox}[Third Prompt]
Critically examine the proof you constructed to identify and resolve any errors or logical gaps. Then, present a complete, rigorous, step-by-step proof. Clearly justify each step with sufficient detail so that an expert can verify your reasoning without needing to fill in any gaps.
\end{promptbox}

Two remarks are in order. First, the theorem stated in the first prompt directly asserts the existence of counterexamples to the Erd\H{o}s--Szemer\'edi sum-product conjecture over $\mathbb R$. If it were unknown whether the conjecture is true or false, this hint could be removed simply by instructing the agent to attempt both a proof and a disproof in separate runs, at the cost of approximately doubling the test-time compute. Second, all problem-specific text is confined to the portion of the first prompt above the separator \texttt{***}. The reasoning instructions---proof-plan proposal, proof construction, and critical review---are problem-agnostic. Consequently, this pipeline can be applied to other theorem-proving tasks by modifying only the problem statement---including any accompanying remark---in the first prompt. For problems that seek a final answer supported by a rigorous argument, minor additional modifications to the prompts are needed.

All trials were conducted via the GPT-5.5 Pro API using the \texttt{gpt-5.5-pro-2026-04-23} snapshot. Tools and web search were disabled. The reasoning effort parameter was set to ``xhigh''. The verbosity parameter was set to ``high'' for the first two rounds, and left at the default (medium) setting for the third round. All other parameters, including temperature and maximum output tokens, were kept at their default values.

\section{Experimental results}

Our agent generated correct proofs in 7 of the 8 independent trials, corresponding to a pass rate of $87.5\%$. The exception was trial 2. In that trial, the argument produced by the model was incomplete after the second round (proof construction) and remained incomplete after the last round (review). The model did not present the incomplete argument as a finished proof. Rather, it emphasized the incompleteness and clearly identified the unresolved gap. For an autonomous mathematical reasoning system, this kind of honest self-diagnosis is desirable when a proof attempt fails.

We report the reasoning-token usage and summary statistics in Table \ref{tab:reasoning_tokens}. Panel (a) gives the round-by-round usage. For each trial, the entry in the \textbf{Total} row is the sum of the reasoning-token counts over the three prompting rounds. Restricting to the seven trials that produced correct proofs, the entries in this row average 125.3k reasoning tokens. Over all eight trials, the entries in the \textbf{Total} row average 132.4k reasoning tokens. Since seven of the eight trials produced correct proofs, the amortized reasoning-token usage per correct proof, counting the failed trial as part of the total compute budget, is 132.4k$\times8/7=$151.3k reasoning tokens.

\begin{table}[!tb]
\centering

\caption{Reasoning-token usage and statistics across the 8 independent trials. Panel (a) reports reasoning-token usage by trial and prompting round. For each trial, the last row, labeled \textbf{Total}, gives the sum of the reasoning-token counts over the three rounds. A checkmark denotes a correct proof, while a dash denotes an incomplete proof. Panel (b) reports the means and sample standard deviations over the seven successful trials, computed for each round and for the \textbf{Total} row in panel (a). Panel (c) reports the corresponding statistics over all eight trials, computed for the \textbf{Total} row in panel (a). Token counts are reported in thousands and rounded to the nearest 0.1k.}
\label{tab:reasoning_tokens}

\vspace{0.25cm}

\textbf{(a) Reasoning-token usage by trial}

\vspace{0.1cm}

\begin{tabular}{lcccccccc}
\toprule
\textbf{Trial no.} & 1 & 2 & 3 & 4 & 5 & 6 & 7 & 8 \\
\midrule
Outcome & \checkmark & -- & \checkmark & \checkmark & \checkmark & \checkmark & \checkmark & \checkmark \\
Round 1: Proof planning & 70.5 & 61.8 & 51.1 & 70.0 & 68.3 & 62.8 & 60.1 & 54.8 \\
Round 2: Proof construction & 71.6 & 64.0 & 31.8 & 61.9 & 25.2 & 29.5 & 57.7 & 39.6 \\
Round 3: Review & 19.8 & 56.4 & 16.1 & 16.5 & 17.6 & 17.4 & 14.0 & 20.6 \\
\midrule
\textbf{Total} & 162.0 & 182.2 & 99.0 & 148.5 & 111.1 & 109.7 & 131.8 & 115.0 \\
\bottomrule
\end{tabular}

\vspace{0.55cm}

\begin{tabular}{@{}c@{\hspace{0.8cm}}c@{}}
\textbf{(b) Statistics over successful trials}
&
\textbf{(c) Statistics over all trials}
\\[0.1cm]

\begin{tabular}{l*{4}{c}}
\toprule
\textbf{Statistic} & \textbf{Round 1} & \textbf{Round 2} & \textbf{Round 3} & \textbf{Total} \\
\midrule
Mean & 62.5 & 45.3 & 17.4 & 125.3 \\
Std. dev. & 7.7 & 18.2 & 2.3 & 23.0 \\
\bottomrule
\end{tabular}
&
\begin{tabular}{lc}
\toprule
\textbf{Statistic} & \textbf{Total} \\
\midrule
Mean & 132.4 \\
Std. dev. & 29.3 \\
\bottomrule
\end{tabular}
\end{tabular}

\end{table}

The first round (proof-plan proposal) was consistently computationally intensive. Across the seven successful trials, it consumed an average of 62.5k reasoning tokens, with a sample standard deviation of 7.7k. The high reasoning-token usage in this round suggests that the model does not settle for a superficial proof plan, but instead invests substantial test-time computation in the central mathematical difficulty of the problem. In this sense, the prompting strategy succeeds in eliciting the desired behavior: the model spends a substantial amount of reasoning effort on finding a proof strategy and developing it in depth.

The reasoning-token usage in the second round is more variable. Among successful trials, this round used an average of 45.3k reasoning tokens, with a much higher sample standard deviation of 18.2k. A natural explanation is that the amount of reasoning required in the second round depends sensitively on the quality of the proof plan produced in the first round. When the first-round plan already contains much of the correct proof strategy, the task in the second round is comparatively light: the model mainly has to fill in details and organize the argument. When the first-round plan points in an unproductive direction, however, reasoning-token usage can be substantially higher, because the model may have to struggle within that approach or abandon it entirely. Trials 1 and 2 have the two largest second-round reasoning token counts, 71.6k and 64.0k, respectively. In both trials, the initial plans went in the wrong direction. In trial 1, however, after extensive reasoning in the second round, the model managed to return to a correct route. In trial 2, by contrast, the model remained stuck and did not complete the proof.

The final round was much lighter for successful trials. Once the second round had already produced a mostly correct proof, the role of the final round was comparatively limited and predictable: check the argument, repair minor issues, and present the final proof cleanly. Accordingly, for the seven successful trials, the final round consumed only 17.4k reasoning tokens on average, with a small sample standard deviation of 2.3k. The unsuccessful trial 2 is again the exception. Since no complete proof had emerged after the second round, the model continued in the final round to try to complete the proof rather than merely polish an existing one, resulting in a much larger final-round usage of 56.4k reasoning tokens.

An analogy is chess. In a winning position, a player who has identified and verified the winning plan can usually make the winning move without much further deliberation; prolonged thinking is often a sign that the plan has not yet been found. Similarly, in our experiments, once the second round has produced a mostly correct proof, the review round is comparatively light. By contrast, unusually high reasoning-token usage in the later rounds indicates that the model is still looking for a route that works or attempting to repair an incomplete one. This suggests a more refined interpretation of test-time computation: reasoning-token consumption records only the amount of computation spent, not how effectively that computation is used. Larger consumption does not necessarily indicate deeper, more comprehensive, or higher-quality reasoning. It may instead reflect uncertainty, detours, or the fact that the model has not yet found an approach it can carry through to the end. Trial 2 is best read in this way: the model had not found a viable proof plan, spent substantial reasoning effort trying to complete and repair the argument, and ultimately reported the remaining gap. Thus the reasoning-token usage data are useful less as a monotone predictor of success than as a diagnostic of where the mathematical difficulty lies and whether the agent has converged to a viable approach.

\section{Proof analysis}

We now compare the following ten proofs: the human proof \cite{BSSZ26}, the OpenAI proof \cite{Ale26}, the Claude Mythos proof \cite{Alp26p}, and seven correct proofs autonomously generated by our GPT-5.5 Pro agent. For these seven, Proof $i$ denotes the final proof produced in trial $i$; in our GitHub repository these files are named \texttt{proof-1.html}, \texttt{proof-3.html}, \texttt{proof-4.html}, \ldots, \texttt{proof-8.html}. The missing index is intentional: trial 2 did not yield a complete proof.

All ten proofs start from the same input from algebraic number theory: an infinite sequence of totally real number fields $K_i$, of degrees $d_i\to\infty$, with uniformly bounded root discriminant. For our purposes, ``totally real'' means that each $K_i$ has $d_i$ real embeddings $\sigma_1,\ldots,\sigma_{d_i}:K_i\to\mathbb R$. Combining these embeddings gives a map
\begin{equation}
        \iota_i:K_i\to \mathbb R^{d_i},\qquad
        \iota_i(x)=(\sigma_1(x),\ldots,\sigma_{d_i}(x)).
\end{equation}
This is the usual Minkowski embedding in the totally real case. Under $\iota_i$, addition and multiplication in $K_i$ correspond to componentwise addition and multiplication in $\mathbb R^{d_i}$. Each $\sigma_j$ preserves sums and products and is injective. Thus the construction can be made inside $K_i$, and applying a chosen $\sigma_j$ gives the desired subset of the real line.

The ten proofs fall naturally into two classes, according to the way they construct the set $A$.

The first class consists of the human proof, the OpenAI proof, the Claude Mythos proof, and Proofs 1, 3, 5, 6, 8. At the level of the guiding idea, these proofs construct $A$ as
\begin{equation}
        A=UP,
\end{equation}
where $U$ is a finite set of units (invertible algebraic integers) and $P$ is a finite set of algebraic integers.

For all proofs in the first class except Proof 6, the bound for the product set comes from the same observation:
\begin{equation} \label{eq:temp}
        AA \subseteq (UU)(PP)\implies|AA|\le |UU|\,|PP|\le |UU|\,|P|^2.
\end{equation}
The set $U$ of units can be chosen so that
\begin{equation}
        |UU|\le C^{d_i}|U|
\end{equation}
for an absolute constant $C>0$. It remains to understand $|A|$. If the multiplication map
\begin{equation}
        U\times P\to K_i,\qquad (u,p)\mapsto up
\end{equation}
is injective, or in some sense close to injective, then
\begin{equation} \label{eq:no}
        |A|\sim|U||P|.
\end{equation}
Together with the estimate (\ref{eq:temp}), we obtain
\begin{equation}
        |AA|\lesssim C^{d_i}|A|^2/|U|.
\end{equation}
Thus it suffices to choose $U$ large enough, compared with the exponential $C^{d_i}$, to obtain an exponential saving in the degree $d_i$. Since $|A|$ grows exponentially in $d_i$, this exponential saving in the $d_i$ translates into a power saving in $|A|$ over the trivial bound $|AA|\le |A|^2$.

The main difference among these proofs lies in how they establish (\ref{eq:no}). Except in Proof 6, one must deal with the possibility that two distinct pairs $(u,p)$ and $(u',p')$ produce the same element, $up=u'p'$. To control such collisions, the human proof and Proof 5 use essentially the same method, whereas the OpenAI proof, the Claude Mythos proof, and Proofs 1, 3, 8 use pairwise distinct methods. Proof 8 is particularly elegant. It obtains the needed injectivity simply by replacing $P$ with a generic translate $q+P$: for all but finitely many integers $q$, no two distinct pairs $(u,p)\in U\times P$ give the same product $u(q+p)$, and hence $|U(q+P)|=|U||q+P|$.

Proof 6 is exceptional within the first class. It still uses the construction $A=UP$, but does not try to prove (\ref{eq:no}). Instead, it uses the trivial lower bound $|A|\ge|P|$. The product set is then controlled by a different argument, not by the estimate (\ref{eq:temp}). The collision problem for the map $(u,p)\mapsto up$ is bypassed rather than solved.

The second class consists of Proofs 4 and 7. These proofs do not use units at all. Instead, they construct $A$ by taking all algebraic integers $x$ such that $\iota_i(x)$ lies in a carefully chosen bounded region of $\mathbb R^{d_i}$. For a small parameter $0<p\le1$, this region has the form
\begin{equation}
        B_{p,i}(T)=
        \left\{
        y=(y_1,\ldots,y_{d_i})\in\mathbb R^{d_i}:
        \sum_{j=1}^{d_i} |y_j|^p \le T^p d_i
        \right\}.
\end{equation}

The regions $B_{p,i}(T)$ behave well under both addition and multiplication. First, since
\begin{equation}
        |x+y|^p\le |x|^p+|y|^p
        \qquad (0<p\le 1),
\end{equation}
one has
\begin{equation}
        B_{p,i}(T)+B_{p,i}(T)\subseteq B_{p,i}(2^{1/p}T).
\end{equation}
This controls the sumset. Second, by Cauchy--Schwarz,
\begin{equation}
        B_{p,i}(T)\cdot B_{p,i}(T)\subseteq B_{p/2,i}(T^2),
\end{equation}
where multiplication is coordinatewise. This controls the product set. Thus, the remaining task is to count, for each of these regions, the algebraic integers $x$ for which $\iota_i(x)$ lies in that region.

One advantage of the second class of proofs is that they require significantly less number-theoretic input. After the common starting point---the existence of totally real fields of growing degree and bounded root discriminant---they do not use units at all. In particular, they do not need Dirichlet's unit theorem or any argument proving that a product map $U\times P\to K_i$ is nearly injective. The construction is also quite canonical: one simply takes the algebraic integers $x$ for which $\iota_i(x)$ lies in a prescribed region. The corresponding counting estimate, however, is technically more involved for these $L^p$-type bodies than for boxes.

\section*{Disclosure of AI assistance}

Beyond using GPT-5.5 Pro for the autonomous agent, GPT-5.5 and GPT-5.5 Pro assisted the author in understanding the proofs generated by the agent. GPT-5.5 Codex assisted in preparing Table~\ref{tab:reasoning_tokens} from
the reasoning-token data. The human author independently verified all proofs and retains full responsibility for the accuracy and integrity of the final content.

\section*{Acknowledgments}

I would like to thank Pengchuan Zhang at OpenAI for providing complimentary access to a ChatGPT Pro account.

\printbibliography

\end{document}